# A Classifier That Teaches Itself: Self-Improving, Frozen-gate Training (SIFT) for Dynamic Document Classification

Bogdan Răduță · Horia Velicu · Alexandru Preda · Șerban Chiricescu
*FlowX.AI*
bogdan.raduta@flowx.ai

**Abstract—** Document classification is a solved problem in the laboratory and an unsolved one in the enterprise. The blocker is rarely model architecture; it is the labeling project that must precede a model and the institutional fear of letting a model retrain itself once one exists. We present **SIFT (Self-Improving, Frozen-gate Training)**, a dynamic classifier service, which attacks both. SIFT serves classification from a deliberately cheap, CPU-bound pipeline, a SPLADE sparse encoder feeding a LightGBM head[5],[6], and escalates only the low-confidence minority of pages to an LLM judge. The judge's verdicts are written back into a labeled corpus, so the expensive model continuously teaches the cheap one: the escalation rate falls, the corpus grows from production traffic rather than from an up-front annotation effort, and accuracy compounds with use. Onboarding a new document family requires only a declarative bundle, label space, anchor phrases, and a judge glossary, not a labeling project. The harder problem is safety: an autonomously retraining classifier can silently regress. SIFT resolves this with a two-part promote gate, a critical-label F1 regression check plus a frozen golden regression set the model is never trained on, either of which vetoes promotion. This turns "retrain monthly without a human" from reckless into routine. We describe the architecture, the self-feeding corpus loop, the frozen-gate promotion mechanism, and an illustrative multi-domain deployment, and we discuss the economics of a classifier whose marginal labeling cost trends toward zero.

## 1 Introduction

Ask why a bank still routes incoming documents by hand, or why an insurer's intake team manually sorts faxes, and the answer is almost never that text classification is technically hard. Supervised classifiers have been reliable for a decade. The answer is that putting one into production requires two things enterprises struggle to supply: a labeled corpus, which means a labeling project with domain experts before a single prediction is served, and a tolerance for retraining, which most organizations lack because a model that retrains itself can silently get worse. The result is a standoff. The model is easy; the labeling project is expensive and slow, and the fear of autonomous drift keeps even a trained model frozen at its launch weights until it quietly decays.

Recent research has supplied the ingredients to break this standoff but not the recipe. Large language models used as judges can label data at a fraction of human cost, with reports of 500-to-5000-fold savings while matching human-to-human agreement[4]; LLM-driven active learning reduces the annotation burden further[3],[9]; and the literature is clear that a small, cheap model trained on good labels often outperforms a large model prompted zero-shot on a narrow domain task. Model cascades, cheap model first and expensive model only when needed, are a well-understood way to control inference cost[7]. Each of these is a paper, not a product, and none of them addresses the second half of the standoff: how to let the resulting classifier retrain itself without a human signing off on every cycle, safely.

> *The bottleneck in production classification is not the model. It is the labeling project that must come before it and the fear of the retraining that must come after. Remove the first and tame the second.*

We present **SIFT (Self-Improving, Frozen-gate Training)**, the dynamic classifier service of the FlowX.AI Platform. SIFT is built on two commitments. The first is that the corpus should assemble itself from production traffic: a cheap model serves the bulk of predictions, an LLM judge handles only the uncertain minority, and the judge's verdicts flow back into the corpus as labeled data, so accuracy compounds with use and a new document family costs a declarative bundle rather than a labeling project. The second is that autonomous retraining must be safe by construction: every candidate model is scored against a frozen golden set it was never trained on and against a per-class regression check, and either can veto promotion. SIFT thus turns the two things enterprises fear, cold-start labeling and self-retraining, into routine, governed operations.

### 1.1 Contributions

- A reframing of the production-classification problem around its real blockers, the up-front labeling project and the unsafety of autonomous retraining, rather than model accuracy.
- A cost-tiered serving architecture (SPLADE sparse features into a LightGBM head, with LLM-judge escalation only on low-confidence pages) whose escalation rate falls as the corpus grows.
- A self-feeding corpus: judge verdicts write back as labeled rows, a bootstrap path labels raw pages with zero data in, and an active-learning queue routes the hardest pages to humans.
- A frozen-gate promotion mechanism, a critical-label F1 regression check plus a never-trained-on golden regression set, that makes autonomous retraining safe.
- Domains expressed as declarative bundles, so a new document family is onboarded without code or a labeling project, with multi-tenant isolation throughout.

## 2 Two Unsolved Problems

### 2.1 The cold-start labeling project

A supervised classifier needs labeled examples, and in a regulated enterprise those labels require domain experts whose time is scarce and expensive. The conventional path, commission a labeling project, wait weeks, then train, makes every new document family a capital expenditure, which is why most never get a model at all. Weak supervision[1] and active learning[2] were designed to reduce this cost, and LLM judges now reduce it dramatically[3],[4]. But a one-time labeling exercise, however cheap, still treats the corpus as a fixed artifact built before deployment, when the most representative labels are precisely the ones production traffic will generate after it.

### 2.2 The unsafety of autonomous retraining

Suppose the labels exist and a model is trained. The model will decay: document mixes shift, new sub-types appear, upstream OCR changes. The remedy is retraining, and the cheap way to retrain is automatically, on a schedule or a trigger. Yet few enterprises permit this, for a sound reason: an automated retrain can regress a class that matters while improving aggregate accuracy, and no one notices until a critical document is misrouted. Concept-drift adaptation is well studied[8], but the operational question, what mechanism makes it safe to promote a freshly trained model without a human reviewing it, is usually answered with “don't,” which is why so many production classifiers are frozen and slowly rotting.

### 2.3 What is needed

Solving one problem without the other is insufficient. Cheap labeling that still produces a frozen model decays; safe retraining that still needs an up-front labeling project never starts. SIFT addresses both at once: a corpus that grows from traffic so labeling is continuous and nearly free, and a frozen evaluation gate so retraining on that corpus is safe to automate. Table 1 contrasts the approaches.

**Table 1. Approaches against the two real blockers.**

| Approach | Cold-start cost | Stays current? | Safe to auto-retrain? |
|---|---|---|---|
| One-time labeling + train | High | No (frozen) | N/A |
| Zero-shot LLM classifier | None | Yes | No model to retrain (slow, costly per call) |
| LLM active-learning label set | Lower | No (still a fixed set) | No explicit gate |
| **SIFT (this work)** | Near zero (bundle) | Yes (traffic-fed) | Yes (frozen gate) |

## 3 The SIFT Architecture

SIFT serves and trains a classifier per document family, called a domain, from a single shared kernel. The serving path is cheap and the expensive judge is the exception, not the rule; the judge's output feeds a corpus that the training path consumes; and a scheduler closes the loop. Figure 1 shows the arrangement.

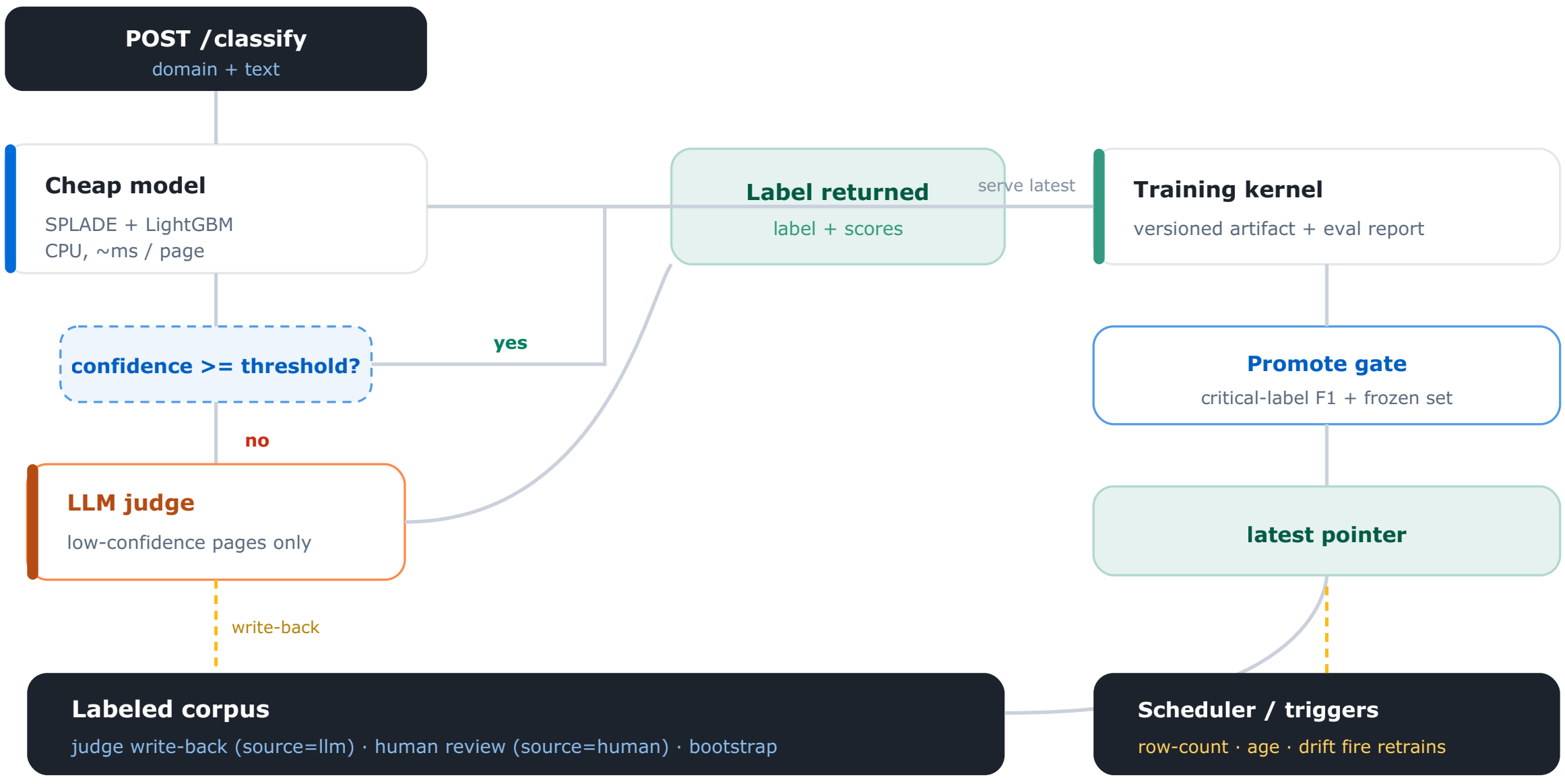

**Figure 1.** The cheap model answers the easy majority on CPU in milliseconds; only low-confidence pages reach the LLM judge. The judge's verdicts, human reviews, and bootstrap labels accumulate in a corpus that the training kernel consumes; a scheduler fires retrains on row-count, age, or drift; and every candidate must clear the promote gate before the latest pointer moves. The expensive model teaches the cheap one, and the escalation rate falls over time.

### 3.1 Domains as data, not code

A domain in SIFT is a declarative bundle: a label space, a few anchor phrases per class for the sparse encoder, optional regex packs, and a glossary and policy block for the judge. The bundle is data, validated by a schema and round-tripping to and from YAML, so a new document family, freight, legal NDAs, healthcare claims, is onboarded by uploading a bundle, with no service redeploy and no per-domain code. One kernel trains and serves them all, and multi-tenant isolation keeps two tenants' identically-named domains, corpora, and model artifacts strictly separate.

### 3.2 Cost-tiered serving

Serving is deliberately cheap. A SPLADE sparse encoder[5] turns each page into lexical features, dotted against per-class anchor phrases and combined with regex hit counts and simple length features; a LightGBM head[6] produces the label and a calibrated confidence, all CPU-bound at milliseconds per page. Only when that confidence falls below a threshold does the page escalate to the LLM judge, which sees the bundle's allowed-label enum, glossary, and policy. This is a classic cost cascade[7], but with a twist that the next subsection makes central: the expensive tier's answers are not discarded after the response, they become training data for the cheap tier.

### 3.3 The self-feeding corpus

Every low-confidence judge verdict that clears a minimum-confidence bar is written back to the corpus as a labeled row, tagged by source. A bootstrap path goes further: pointed at a pile of raw, unlabeled pages, it runs the judge over each with the bundle's allowed labels and persists the confident verdicts, producing a first corpus and a first model from zero labeled data in. Human reviewers enter through an active-learning queue that surfaces the least-confident judge-labeled rows first, so scarce expert attention lands on the hardest pages and their corrected labels are stamped as ground truth. The corpus therefore grows fastest exactly where the model is weakest, and because the judge handles a shrinking fraction of traffic as the cheap model learns, the marginal cost of a new labeled example trends toward zero.

## 4 Safe Autonomous Retraining

A corpus that grows from traffic is only useful if the model can retrain on it without a human in every loop, and that is safe only if a bad candidate cannot be promoted. SIFT's promote gate is the mechanism that earns the autonomy. Figure 2 shows its logic.

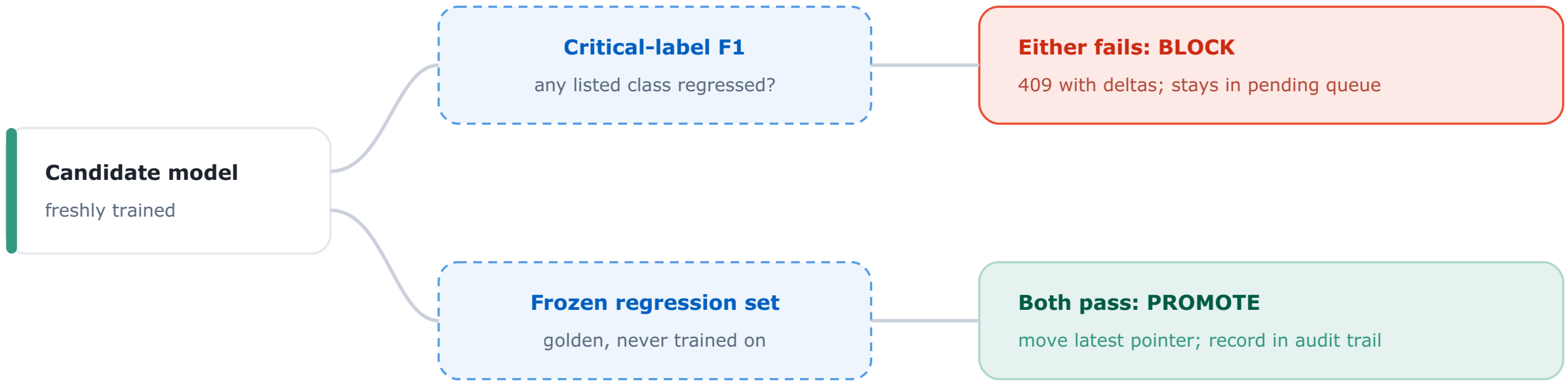

**Figure 2.** A candidate must clear both checks. The bundle's critical_labels name the classes that must never regress; if any candidate F1 falls below the current model's, promotion is blocked with the regression deltas. Independently, the candidate is scored against a frozen golden set carried in the bundle and never used for training; a drop there also vetoes promotion. Auto-triggered retrains land in a pending-review queue, and only a clean pass moves the live pointer.

### 4.1 Critical labels that refuse to regress

Not every class matters equally. A bundle declares a list of critical labels, the classes whose misclassification is expensive, and the gate refuses to promote any candidate whose F1 on a critical label is below the current model's. Where there are no critical labels, the gate is permissive and any successful train promotes; where they are declared, a candidate with no evaluation is rejected outright, and a candidate that regresses a critical class is blocked with explicit per-label deltas. The protection is targeted: aggregate accuracy can fluctuate freely, but the classes the business cannot afford to get wrong are held to a ratchet.

### 4.2 A frozen set the model never sees

The second check guards against the subtle failure where a candidate looks better on data drawn from the same shifting distribution that produced it. The bundle carries a frozen regression set, a small collection of golden examples that is never added to the training corpus, and every candidate is scored against it. A drop on the frozen set, on critical-label F1 or overall accuracy, is a hard veto. Because the set is immutable and held out by construction, it is a stable yardstick across retrains: improvement measured against it is real, not an artifact of the corpus drifting underneath the metric.

### 4.3 Why the gate changes the economics of retraining

With these two checks in place, the calculus of autonomy inverts. A scheduler can fire retrains on row-count, age, or drift triggers; auto-triggered candidates land in a pending queue and either auto-promote when the gate is clean and the configured accuracy improvement is met, or wait for a human when it is not. The downside risk of an automated retrain, a silently regressed critical class, is the precise thing the gate forecloses. "Retrain this classifier every month without asking anyone" stops being a sentence that alarms a risk officer and becomes a configuration default, which is what lets the self-feeding corpus actually translate into a continuously improving production model.

## 5 Illustrative Application

Consider an operations team that must classify the pages of incoming legal documents, separating an actual non-disclosure agreement (the critical class, NDA) from pages that merely reference one (SUPPORTING) and from everything else (MISC). The figures here are illustrative and chosen to expose the mechanism.

There is no labeling project. The team authors a bundle: three labels, a handful of anchor phrases ("Non-Disclosure Agreement," "Confidentiality Agreement"), a regex for the literal heading, a judge policy clarifying that a page titled "Non-Disclosure Agreement" is an NDA while a passing mention is SUPPORTING, and NDA as the sole critical label. They point the bootstrap path at a few hundred unlabeled pages; the judge labels the confident ones, a first corpus forms, and a first model trains, all without a human labeling a single page. The eval report, a confusion matrix, per-class precision and recall, and a confidence histogram, is generated automatically for review.

In production the cheap model handles the clear majority of pages in milliseconds; on a well-specified domain roughly fifteen to twenty-five percent of pages fall below the confidence threshold and escalate to the judge, each costing a fraction of a cent. Those judge verdicts flow back into the corpus. A month later the scheduler fires a retrain on the grown corpus and produces a candidate that improves overall accuracy but, the gate discovers, drops NDA F1 by two points: promotion is blocked with the per-label delta and the candidate waits in the pending queue for a human, exactly as intended. The team inspects, finds a cluster of mislabeled SUPPORTING pages from an ambiguous template, corrects them through the active-learning queue, and the next candidate clears both the critical-label check and the frozen golden set and auto-promotes. Over successive cycles the escalation rate falls as the cheap model absorbs what the judge taught it, and the team's only recurring cost is reviewing the occasional blocked promotion. The classifier got better while nobody ran a labeling project, and it never once shipped a regression on the class that mattered.

## 6 Empirical Evaluation

Section 5's figures are explicitly illustrative. To test the actual mechanism, the cost-tiered cascade of Section 3.2 and the promotion gate of Section 4, we built a standalone reference implementation (a TF-IDF+LightGBM classifier standing in for the paper's SPLADE+LightGBM combination) and ran it on AG News, a public four-class news-topic benchmark, with "Business" standing in for a compliance-critical label such as the NDA example in Section 5. The judge role was filled independently by two model providers (Claude Haiku 4.5, GPT-5.4-mini), so the results below are not an artifact of one model's behavior. The implementation, raw results, and a pytest suite are released as supplementary material.

### 6.1 Setup

A deliberately weak cold start (300 seed documents per class, 1,200 total) produced a seed classifier at 66.1% accuracy, well below what more seed data would give, so the cascade has real work to do. At a fixed confidence threshold, 40.4% of a frozen 1,000-document golden set and 41.0% of an 800-document write-back pool fell below the threshold and escalated, somewhat above the "fifteen to twenty-five percent" the illustrative section calls typical for a well-specified domain, which is expected: a weaker seed model escalates more.

### 6.2 The cascade lifts accuracy, and write-back improves the classifier itself

Table 1 reports both halves of the mechanism: the inference-time cascade (classifier answer, judge answer on low-confidence rows) against the classifier alone, and a second, cheap classifier retrained on the seed plus the judge's labels for the escalated write-back pool, evaluated *without* any escalation at all, isolating whether the write-back loop makes the classifier itself better.

**Table 1. Cascade and write-back results on the frozen golden set (n=1,000), classifier-alone baseline = 66.1% accuracy / 0.609 Business-F1.**

| Judge | Cascade accuracy | Judge accuracy on escalated rows | Round-1 accuracy (no escalation) | Round-1 Business-F1 |
|---|---|---|---|---|
| Claude Haiku 4.5 | 82.8% | 83.2% | 71.8% | 0.694 |
| GPT-5.4-mini | 83.3% | 84.4% | 71.8% | 0.693 |

Both judges lift cascade accuracy from 66.1% to over 83%, and, separately, retraining on the judge-labeled write-back pool alone (no escalation at inference time) lifts the classifier's own accuracy to 71.8%, a real gain purely from the self-feeding corpus, with Business-F1 improving from 0.609 to ~0.69 rather than regressing: the promote-gate correctly cleared both candidates.

### 6.3 Does the gate actually gate?

A promote-gate that always says yes is not a gate. We tested the rejection path directly: a third candidate was trained on the same seed and pool, but with a deliberately corrupted batch (60% of Business-labeled documents in both the seed and the escalated pool relabeled to a wrong class, simulating a systematic labeling bug), with no judge model involved. The corrupted candidate's Business-F1 collapsed to 0.345, a clear regression from the 0.609 baseline, and the promote-gate correctly **blocked** it. Section 6.2's two genuine improvements were promoted and this deliberate regression was not, which is the complete claim: the gate passes real gains and blocks real regressions, not just the former.

## 6.4 What this evaluation does and does not show

This tests the cascade and the critical-label half of the promotion gate (Section 4.1) on a standalone classifier and a public benchmark; it does not exercise the frozen-golden-set-on-a-shifting-corpus check specifically (both round-1 evaluations here use the same frozen golden set the paper describes, but we did not test corpus drift over time), active-learning review, multi-tenant isolation, or the domain-bundle authoring flow. The result supports the paper's central economic claim, cascade plus write-back materially improves a weak cold-start classifier at a fraction of full-judge cost, and, new information beyond what the paper argued, demonstrates the gate's rejection path with a genuine adversarial candidate rather than only its acceptance path. The reference implementation, the raw judge sweep, and a pytest suite that reproduces Table 1 and the adversarial check are released as supplementary material.

## 6.5 The escalation judge need not be a frontier model

Sections 6.1–6.4 use closed frontier models (Claude Haiku 4.5, GPT-5.4-mini) as the judge. A follow-up run asks two further questions: does the harness still work when the judge is an open-weight model an operator can host themselves, and does fine-tuning that judge on the operator's own traffic close the remaining gap to a frontier judge. This reuses the same frozen 1,000-item golden set and the same TF-IDF+LightGBM cheap tier, now trained on the full AG News training split (steady-state) rather than the deliberately weak 300-per-class cold start of Sections 6.1–6.4, since a cold-start classifier escalates too much of its traffic for a judge-quality comparison to be informative.

Used alone with no cascade, an open-weight model (Qwen3.5 9B, served locally) trails the strongest tested frontier model (Claude Opus 5) by 11.7 points, 80.6% against 92.3%. Routed through the harness, so that only the classifier's genuinely uncertain minority reaches the open-weight judge, most of that gap closes: 90.9% accuracy at a 21.3% escalation rate, rising to 91.7% once the confidence threshold is retuned for the steady-state classifier, at roughly 1/3,400th the per-task cost of calling the frontier model directly. At this point the harness still trails the frontier model by 0.6 points. Extending the frontier side to five further widely used models, each classifying the same golden set directly with no cascade, Claude Haiku 4.5, GPT-5.6 "Luna" and "Terra" configurations, Grok 4.6, and Gemini 3.5 Flash, scores them between 85.0% and 92.1%, showing the open-weight harness result is not an artifact of Claude Opus 5 specifically.

Table 2 adds a fifth condition: fine-tuning the open-weight judge itself, a LoRA adapter trained on the same corpus the cheap classifier learned from. The tuned judge alone reaches 91.1% accuracy, up from 80.6% untuned, and behind the harness it reaches 92.5% at a 7.3% escalation rate and roughly 1/4,075th the frontier model's per-task cost. That is a narrow edge over the two strongest frontier baselines (Claude Opus 5, 92.3%; Gemini 3.5 Flash, 92.1%) and a clear one over the other four (2.5 to 7.5 points: Grok 4.6, GPT-5.6 Terra, GPT-5.6 Luna, Claude Haiku 4.5). Given how narrow the margin is over the single strongest baseline, we would not describe this as the harness beating frontier intelligence outright, only as closing the gap to it while removing nearly all of its cost, which is the more defensible and more useful claim of the two.

**Table 2. Judge-model comparison on the frozen golden set (n=1,000), steady-state cheap classifier (91.0% alone). Cost is real measured token usage × real published per-token pricing (Anthropic list price for Claude; Artificial Analysis hosted API rate for Qwen3.5 9B, retrieved 2026-08-27).**

| Condition | Judge model | Accuracy | Cost / task (USD) | Escalation rate |
|---|---|---|---|---|
| Open-weight judge alone | Qwen3.5 9B | 80.6% | 0.0000173 | n/a |
| Harness, untuned judge (retuned threshold) | Qwen3.5 9B | 91.7% | 0.0000012 | 7.3% |
| Fine-tuned open-weight judge alone | LoRA Qwen3.5 9B | 91.1% | 0.0000146 | n/a |
| Harness, fine-tuned judge | LoRA Qwen3.5 9B | **92.5%** | 0.0000010 | 7.3% |
| Frontier model alone | Claude Opus 5 | 92.3% | 0.0041018 | n/a |

This extension carries its own scope limits: it was run on one hardware and serving configuration (Apple-silicon MLX for the fine-tune), the fine-tune was trained once rather than across multiple seeds, and "closes the gap to frontier" is a comparison against a single snapshot of each vendor's model rather than a moving target. It is released, like Sections 6.1–6.4, as supplementary material with a pytest suite that checks these claims against frozen result files.

## 7 Discussion

The economic shape of SIFT is its point. A conventional classifier front-loads cost (the labeling project) and then decays; SIFT amortizes labeling across production traffic at a marginal cost that falls as the cheap model learns, and the frozen gate converts the recurring cost of staying current from "a labeling project every quarter" into "reviewing the occasional vetoed promotion." The LLM judge is best understood not as the classifier but as a teacher hired by the hour, deployed only on the lessons the student has not yet learned, and progressively idled as the student improves. This is a different use of an expensive model than running it on every request, and a far cheaper one at steady state. Section 6.5 makes the idling literal: once the teacher's own verdicts have trained a judge an operator can host, the harness no longer needs to rent frontier intelligence at all to hold its accuracy.

SIFT also slots into a broader platform pattern shared with related systems. Its eval-gated promotion mirrors the offline-then-online discipline used for hallucination control, its judge-write-back loop is a close cousin of recursive self-improvement, and its promotions, audit trail, and per-tenant isolation make it governable by the same policy engine that governs every other agent. A classifier that teaches itself is only deployable in a regulated enterprise because the teaching is gated, audited, and scoped, which is the same reason the other components in the platform are.

## 8 Limitations

- **The empirical evaluation uses a stand-in classifier and domain.** Section 6 tests the cascade and the critical-label promote-gate on TF-IDF+LightGBM and AG News, not SPLADE and a document-family bundle; it does not exercise active-learning review, multi-tenant isolation, or the bundle-authoring flow, and its escalation rate (~40%) is higher than the "fifteen to twenty-five percent" Section 5 calls typical, because the seed classifier was deliberately weakened to stress-test the cascade.
- **Text only.** SIFT classifies OCR-derived text; it does not yet use image embeddings, so a domain that depends on visual layout rather than words is out of scope until multimodal features land.
- **The judge can be wrong.** Write-back trusts the judge above a confidence bar, so a systematically biased judge can teach the cheap model its bias. The frozen set and active-learning review are the checks, but a blind spot shared by judge and golden set would persist.

- **The frozen set must be representative.** A golden regression set that does not cover an emerging sub-type cannot veto a regression on it. The set is a fixed yardstick, which is its strength and, against genuinely novel inputs, its limit.
- **Cold-start quality depends on the bundle.** Bootstrap quality rests on the anchor phrases, glossary, and policy an author supplies; a thin bundle yields a noisy first corpus that takes longer to converge.
- **Operational footprint.** The sparse encoder is the dominant resident cost, and training currently runs in-process unless the queue mode is enabled; large multi-replica deployments need the shared artifact store and queue that are on the roadmap.
- **The open-weight judge result is a single snapshot.** Section 6.5's fine-tuned harness was trained once, on one hardware and serving configuration, and its edge over the strongest frontier baseline is narrow (0.2 points) and measured against one snapshot of each vendor's model rather than a moving target; it should be read as evidence that a self-hosted judge can close most of the gap to frontier accuracy, not as a durable ranking.

## 9 Conclusion

Production text classification has been held back not by models but by the labeling project that precedes them and the fear of the retraining that should follow. SIFT removes the first by letting the corpus assemble itself from production traffic, with an LLM judge teaching a cheap model on exactly the pages it finds hard, and tames the second with a frozen evaluation gate that makes autonomous retraining safe by refusing to promote a candidate that regresses a critical class or a held-out golden set. The result is a classifier whose accuracy compounds with use, whose marginal labeling cost trends toward zero, and whose self-improvement is governed rather than reckless. Onboarding a new document family becomes a bundle, not a project, and keeping it accurate becomes a default, not a campaign.

## References

**[1]** Ratner, A., Bach, S. H., Ehrenberg, H., Fries, J., Wu, S., & Ré, C. *Snorkel: Rapid Training Data Creation with Weak Supervision.* Proc. VLDB Endowment, 11(3):269–282, 2017.

**[2]** Settles, B. *Active Learning Literature Survey.* University of Wisconsin–Madison, Computer Sciences Technical Report 1648, 2009.

**[3]** Rouzegar, H., & Makrehchi, M. *Enhancing Text Classification through LLM-Driven Active Learning and Human Annotation.* Proc. LAW 2024. arXiv:2406.12114.

**[4]** Zheng, L., Chiang, W.-L., Sheng, Y., et al. *Judging LLM-as-a-Judge with MT-Bench and Chatbot Arena.* Advances in Neural Information Processing Systems (NeurIPS) 36, 2023.

**[5]** Formal, T., Piwowarski, B., & Clinchant, S. *SPLADE: Sparse Lexical and Expansion Model for First Stage Ranking.* Proc. SIGIR 2021.

**[6]** Ke, G., Meng, Q., Finley, T., et al. *LightGBM: A Highly Efficient Gradient Boosting Decision Tree.* Advances in Neural Information Processing Systems (NeurIPS) 30, 2017.

**[7]** Chen, L., Zaharia, M., & Zou, J. *FrugalGPT: How to Use Large Language Models While Reducing Cost and Improving Performance.* arXiv:2305.05176, 2023.

**[8]** Gama, J., Žliobaitė, I., Bifet, A., Pechenizkiy, M., & Bouchachia, A. *A Survey on Concept Drift Adaptation.* ACM Computing Surveys, 46(4):1–37, 2014.

**[9]** Zhang, Y., & Takada, S. *Applying LLMs to Active Learning: Toward Cost-Efficient Cross-Task Text Classification Without Manually Labeled Data.* International Journal of Intelligent Systems, 2025. arXiv:2502.16892.

**[10]** National Institute of Standards and Technology. *Artificial Intelligence Risk Management Framework (AI RMF 1.0).* NIST AI 100-1, 2023.

## Appendix A  Anatomy of a Domain Bundle

A domain is data. The bundle below (abbreviated) is everything needed to onboard a new document family: no code, no labeling project.

```
name: legal-nda
version: "1"
critical_labels: [NDA]          # classes that must never regress

judge_policy:
  extra_instructions: |
    A page headed "Non-Disclosure Agreement" is NDA.
    Pages that mention NDAs in passing are SUPPORTING.

labels:
  - name: NDA
    pos_queries: ["Non-Disclosure Agreement", "Confidentiality Agreement"]
    regex_pos: ["\bnon[-\s]disclosure\s+agreement\b"]
  - name: SUPPORTING
  - name: MISC

regression_set:                 # frozen golden examples, never trained on
  - {page_id: g-1, label: NDA, text: "MUTUAL NON-DISCLOSURE ..."}

triggers: { enabled: true, min_new_rows_since_last_train: 200, drift_threshold: 0.08 }
auto_promote: { enabled: true, require_accuracy_improvement: true }
```

## Appendix B  The Continuous-Training Loop

The five steps below run unattended once configured; the gate of Section 4 is what makes step five safe to automate.

| Step | What happens | Governing knob |
|---|---|---|
| 1. Judge write-through | Low-confidence /classify verdicts persist as labeled rows (source=llm) | autolabel min-confidence |
| 2. Trigger scheduler | Walks each domain; row-count / age / drift fire a retrain; in-flight jobs block stacking | triggers block in bundle |
| 3. Pending queue | Auto-triggered trains land as candidates; one pending per domain (supersede) | auto_promote settings |
| 4. Frozen gate | Critical-label F1 + golden-set score; either drop vetoes promotion | critical_labels, regression_set |
| 5. Drift detection | Rolling confidence mean vs training baseline triggers the next cycle | drift_threshold, window |
| Active learning | Least-confident rows surfaced for human approve / relabel / reject | review queue limit |